\documentclass[nohyperref,accepted]{article}

\usepackage{microtype}
\usepackage{graphicx}
\usepackage{booktabs} % for professional tables
\usepackage{wrapfig}
\usepackage{hyperref}

\usepackage{icml2022}

\usepackage{tabularx}
\usepackage{multirow}
\usepackage{subcaption}

\usepackage{amsmath}
\usepackage{amssymb}
\usepackage{mathtools}
\usepackage{amsthm}
\usepackage{amsfonts}

\usepackage[capitalize,noabbrev]{cleveref}

\usepackage{pdflscape}
\theoremstyle{plain}

\theoremstyle{definition}

\theoremstyle{remark}

\usepackage[textsize=tiny]{todonotes}

\icmltitlerunning{Adiabatic Quantum Support Vector Machines}

\begin{document}

\twocolumn[
\icmltitle{Adiabatic Quantum Support Vector Machines}



\icmlsetsymbol{equal}{*}

\begin{icmlauthorlist}
\icmlauthor{Prasanna Date}{ornl}
\icmlauthor{Dong Jun Woun}{utk}
\icmlauthor{Kathleen Hamilton}{ornl}
\icmlauthor{Eduardo A. Coello Perez}{ornl}
\icmlauthor{Mayanka Chandra Shekhar}{ornl}
\icmlauthor{Francisco Rios}{ornl}
\icmlauthor{John Gounley}{ornl}
\icmlauthor{In-Saeng Suh}{ornl}
\icmlauthor{Travis Humble}{ornl}
\icmlauthor{Georgia Tourassi}{ornl}
\end{icmlauthorlist}

\icmlaffiliation{ornl}{Oak Ridge National Laboratory, Oak Ridge, TN, United States}
\icmlaffiliation{utk}{University of Tennessee, Knoxville, TN, United States}
 
\icmlcorrespondingauthor{Prasanna Date}{datepa@ornl.gov}

\icmlkeywords{Quantum Machine Learning, Quantum AI, Support Vector Machine, Machine Learning, Quantum Computing}

\vskip 0.3in
]




\begin{abstract}
Adiabatic quantum computers can solve difficult optimization problems (e.g., the quadratic unconstrained binary optimization problem), and they seem well suited to train machine learning models. In this paper, we describe an adiabatic quantum approach for training support vector machines. We show that the time complexity of our quantum approach is an order of magnitude better than the classical approach. Next, we compare the test accuracy of our quantum approach against a classical approach that uses the Scikit-learn library in Python across five benchmark datasets (Iris, Wisconsin Breast Cancer (WBC), Wine, Digits, and Lambeq). We show that our quantum approach obtains accuracies on par with the classical approach. Finally, we perform a scalability study in which we compute the total training times of the quantum approach and the classical approach with increasing number of features and number of data points in the training dataset. Our scalability results show that the quantum approach obtains a 3.5--4.5$\times$ speedup over the classical approach on datasets with many (millions of) features.
\end{abstract}

\section{Introduction}
\label{sec:intro}
In the 21st century, computer science has witnessed the rise of machine learning technologies and an explosion of their application areas \cite{lecun2015deep}.
Machine learning applications now run on all devices---ranging from edge devices such as smartphones to large supercomputing systems \cite{date2019combinatorial}.
When developing any end-to-end machine learning application, training the machine learning model takes a significant amount of time and is a major bottleneck \cite{munson2012study}.
Training a machine learning model generally refers to obtaining a set of optimal learning parameters that minimize a well-defined error function (i.e., an optimization problem) \cite{abu2012learning}.

Although algorithms for solving optimization problems exist on classical computers, quantum computers are thought to be better at solving them \cite{moll2018quantum}.
Interest in quantum computing has been growing rapidly in recent years as Moore's law reaches its inevitable conclusion---a scenario in which classical computing can no longer sustain exponential leaps in performance~\cite{theis2017end}.
% Quantum computing uses quantum bits (qubits) as fundamental units of computation, and leverages quantum mechanical phenomena like superposition, entanglement and tunneling to perform computation.
% Two quantum computing paradigms exist in the present day: Adiabatic Quantum Computing (AQC), which leverages quantum annealing to find the ground state of Ising Hamiltonians; and Universal Quantum Computing (UQC), which leverages a gate-based model to control inter-qubit interactions.
For certain problems, some quantum algorithms
% use hardware that is sensitive to noise, and are said to be in the Noisy Intermediate-Scale Quantum (NISQ) era of quantum computing. 
% Despite being noisy, these NISQ machines 
outperform their classical contemporaries, including the Fourier transformation \cite{coppersmith2002approximate}, integer factorization \cite{shor1994algorithms}, and database search \cite{grover1996fast}, and have attained \textit{quantum supremacy} \cite{arute2019quantum}.
% This provides a strong motivation for exploring machine learning on quantum computers.
Developing machine learning algorithms for quantum computing should result in shorter training times \cite{perdomo2018opportunities}, but this approach must be tested thoroughly.

To this end, we quantify the gains obtained from using an adiabatic quantum computer (AQC) to train support vector machines (SVM).
AQCs are adept at approximately solving the quadratic unconstrained binary optimization (QUBO) problem, which is known to be NP-hard \cite{kochenberger2014unconstrained}.
In this paper, we formulate SVM training as a QUBO problem and use the D-Wave Advantage AQC to solve it.
% We specifically pick SVMs because their decision functions are linear, and error functions are quadratic, allowing them to be formulated as QUBO problems and solved on an AQC.
% Our results indicate that the theoretical time complexity of training SVM models using our quantum approach is $\mathcal{O}(N^2)$, which is an order of magnitude better than the conventional classical approach ($\mathcal{O}(N^3)$).
% \textbf{We also test our quantum approach on the ...}
The main contributions of this work are as follows:
\begin{enumerate}
    \item We formulate an adiabatic quantum approach for training SVMs. 
    \item We show that the time complexity of our quantum approach is an order of magnitude faster than the current classical approach.
    \item We compare the test accuracies of our quantum approach to those of the classical approach across five benchmark datasets: Iris, Wisconsin Breast Cancer (WBC), Wine, Digits, and Lambeq. Our results show that the quantum approach obtains accuracies on par with the classical approach.
    % \item We conducted empirical tests to evaluate the performance of our quantum approach on the D-Wave Advantage adiabatic quantum annealer, as well as a classical approach using the Scikit-learn library in Python on x86-64 instruction set processors. In order to compare the two approaches, we focused on accuracy and computation time as the primary performance metrics. To train and test our models, we utilized various datasets, such as Iris, Wine, Digits, Wisconsin Breast Cancer, and Synthetic, Lambeq datasets.
    \item We show that the quantum approach is 3.5--4.5$\times$ faster than the classical approach for training SVMs on large (millions of features) synthetic datasets. 
    % \item We demonstrate that our quantum approach achieves comparable accuracy to the classical approach. Furthermore, we observed the quantum approach achieving a $3.69\times$ speedup over the classical approach on larger datasets.
\end{enumerate}

\section{Related Work}
\label{sec:related}
The incorporation of quantum information and quantum computing is expected to have a significant effect on  machine learning \cite{pudenz2013quantum,dunjko2016quantum,biamonte2017quantum}.  
A hallmark of quantum algorithms is leveraging physical phenomena such as superposition and entanglement in addition to an exponentially large Hilbert space for computational advantages \cite{humble2022snowmass,delgado2022quantum}.  
% There are many examples of algorithms that predict quantum advantage, but many of these algorithms require either fault-tolerant quantum systems, or quantum registers that far exceed the size of current quantum hardware.  
Two main areas of research in quantum machine learning are the development of quantum algorithms that can speed up the training of classical models and the development of quantum models that act as parameterized learning models \cite{dunjko2018machine,benedetti2019parameterized}.

Quantum machine learning has been explored primarily on universal quantum computers.
% Reducing the computational overhead associated with classical training: one example is the applications of the 
For example, the Harrow-Hassim-Lloyd algorithm \cite{harrow2009quantum} is used to speed up matrix inversion.  
Jordan's algorithm \cite{jordan2005fast} was adapted by Gilyen et al.~\cite{gilyen2019optimizing} to leverage the quantum Fourier transform to compute the gradient of a classical function with a sublinear number of steps. 
Date proposed the quantum discriminator, which is a quantum discriminant model used for binary classification \cite{date2024quantum}.
Quiroga et al. propose the quantum k-means classifier on the IBM quantum computers for discriminating quantum states on hardware \cite{quiroga2021discriminating}.

Quantum gradient descent methods have also seen development \cite{rebentrost2019quantum}.  
Furthermore, the general training of machine learning classifiers has been studied, and these investigations show that sublinear training is possible \cite{pmlr-v97-li19b}.  
% the authors derive bounds on the training time using a quantum oracle, and 
However, these algorithms require either fault-tolerant quantum systems or quantum registers that far exceed current quantum hardware.
Li et al. propose the ST-VQC algorithm to integrate non-linearity in quantum learning and improve the robustness of the learning model to noise. 
The number of qubits available on near-term, gate-based hardware has limited the direct application of quantum computing to standard machine learning benchmarks.  Quantum classifiers that can analyze standard benchmark datasets (e.g., Wine, Sonar) have been studied via numerical simulations \cite{schuld2018circuit}, but studies that use hardware either require compression methods to reduce the number of training features to a tractable size or are limited to a small number of features \cite{havlivcek2019supervised}.  The size of qubit registers continues to increase, and the quality of gates continues to improve; as a result, exploring the applications of parameterized circuit models remains an active area of research \cite{benedetti2019parameterized}.

Instead of implementing a quantum algorithm as a series of unitary gates applied to a qubit register, the D-Wave processor encodes a problem into a system of interacting quantum spins. Through a gradual change in the system's Hamiltonian, the process of quantum annealing is used to find the ground state (and correspondingly the solution) of the encoded problem.   Casting machine learning tasks as AQC problems through the use of QUBOs is different from constructing a quantum circuit \cite{pudenz2013quantum}.
But, because of the similarities between QUBOs and Ising spin models \cite{lucas2014ising}, several QUBO-based analogues of restricted Boltamann machines, deep belief networks, and Hopfield networks have been proposed in the literature \cite{amin2018quantum,date2019classical}.
Chen et al. use the D-Wave system to solve an NP-hard problem that pertains to energy-efficient routing in wireless sensor networks.
QUBO-based implementations of conventional machine learning models (e.g., linear regression, k-means clustering, SVMs) have also been developed for the D-Wave platform \cite{amin2018quantum,date2019classical,arthur2021balanced,date2021adiabatic,date2021qubo}.  
The strength of quantum annealing lies in the ability to solve difficult optimization problems, and this ability means the D-Wave platform could speed up training for classical machine learning models \cite{qboost_2012neven,adachi2015application,willsch2019support}.

Recent efforts have explored quantum approaches for SVMs as well. 
Otgonbaatar successfully demonstrated training SVMs on large remote sensing data by using the D-Wave quantum annealer \cite{9884273}. To train large datasets, his team employed corsets, which are smaller representative subsets of the data. Barbosa analyzed the factors that contribute to the difficulty of solving a Maximum Clique problem on the D-Wave quantum computer \cite{a14060187}. Lee investigated more effective ways to formulate QUBO problems for linear systems on the D-Wave AQCs \cite{10.1117/12.2632416}, and Simoes's research shows that quantum SVMs and neural networks trained on universal quantum computers can achieve higher accuracy than classical approaches \cite{10015720}.

\section{SVMs}
\label{sec:svm}

We use the following notation throughout this paper:
\begin{itemize}
    \item $\mathbb{R}$, $\mathbb{N}$, $\mathbb{B}$: Set of real, natural, and binary numbers, respectively.
    % \item $\mathbb{B}$: Set of binary numbers, i.e. $\mathbb{B} = \{0, 1\}$.
    % \item $\mathbb{N}$: Set of natural numbers.
    \item $X$: Training dataset, $X \in \mathbb{R}^{N \times d}$; $N, d \in \mathbb{N}$.
    \item $Y$: Labels for binary classification; $y_i$ is $+1$ ($-1$) if data point $x_i \in X$ belongs to the first (second) class.
    \item $w$, $b$: Weights and bias of the SVM; $w \in \mathbb{R}^d$, $b \in \mathbb{R}$.
\end{itemize}

\begin{figure}
    \centering
    \includegraphics[width=0.48\textwidth]{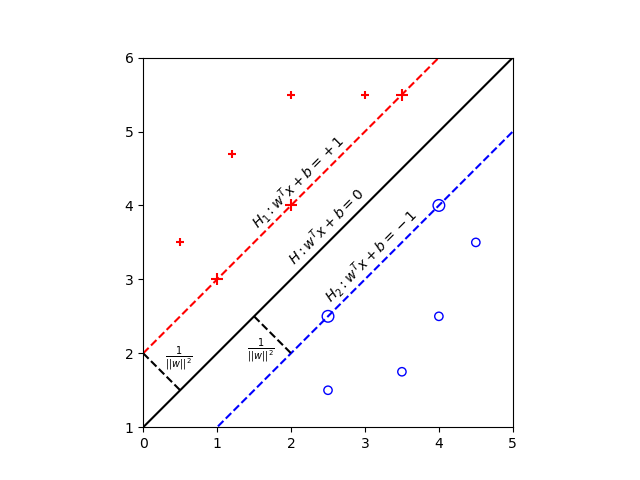}
    \caption{SVMs.}
    \label{fig:svm-explained}
\end{figure}

We now state the SVM training problem shown in Figure \ref{fig:svm-explained}, where we have two classes of data shown by red pluses and blue circles.
We would like to find a separating hyperplane, $H: w^T x + b = 0$, that maximizes the distance between the nearest points that belong to the two classes.
Hyperplanes $H_1: w^T x + b = +1$ and $H_2: w^T x + b = -1$ are parallel to the separating hyperplane $H$ and demarcate the boundary of the corridor in which $H$ lies.
$H_1$ and $H_2$ can be thought of as the hyperplanes in which the first ($H_1$) and second ($H_2$) classes  begin.
The orthogonal distance between $H_1$ and $H_2$ is given by $\frac{2}{||w||^2}$, which we would like to maximize.
This is equivalent to minimizing $||w||^2$.

We must also ensure that all the points are classified correctly.
All the red pluses in Figure \ref{fig:svm-explained} belong to the first class (i.e., $y_i = +1$).
All the blue circles in Figure \ref{fig:svm-explained} belong to the second class (i.e., $y_i = -1$).
We must ensure that we have $w^T x_i + b \ge 1$ for all the points in the first class and $w^T x_i + b \le -1$ for all the points in the second class.
By incorporating these constraints for all points in the training dataset, the SVM training problem can be stated as follows:
\begin{align}
    & \min_{w, b} \ ||w||^2 \label{eq:svm-qpp} \\
    \text{subject to:} \quad & y_i (w^T x_i + b) \ge 1 \qquad \forall i = 1, 2, \ldots, N \nonumber
\end{align}

% This is a quadratic programming problem, where the objective function is quadratic and the constraints are linear in weights $w$ and bias $b$.
The objective function is convex because its Hessian is positive semi-definite.
Furthermore, the constraints are linear, and hence, convex. 
Problem \ref{eq:svm-qpp} is therefore a quadratic programming problem.
To solve Problem \ref{eq:svm-qpp}, we first compute the Lagrangian dual as follows:
\begin{align}
    \max_{w, b, \lambda} \ \mathcal{L}(w, b, \lambda) = ||w||^2 - \sum_{i = 1}^{N} \lambda_i \left[ y_i (w^T x_i + b) - 1 \right], \label{eq:lagrangian-dual}
\end{align}

where, $\lambda$ is the vector that contains all the Lagrangian multipliers (i.e., $\lambda = [\lambda_1 \ \lambda_2 \ \cdots \ \lambda_N]^T$, and $\lambda_i \ge 0 \ \forall i$). 
The non-zero Lagrangian multipliers in the final solution correspond to the support vectors and determine the hyperplanes $H_1$ and $H_2$ in Figure \ref{fig:svm-explained}.
The Lagrangian dual problem (Eq.~\ref{eq:lagrangian-dual}) is solved in $\mathcal{O}(N^3)$ time on classical computers by applying the Karush-Kuhn-Tucker (KKT) conditions \cite{karush1939minima, kuhn2014nonlinear}. 
As part of the KKT conditions, we set the gradient of $\mathcal{L}(w, b, \lambda)$ with respect to $w$ to 0. 
We also set the partial derivative of $\mathcal{L}(w, b, \lambda)$ with respect to $b$ to zero.
Doing so yields the following:
\begin{align}
    \nabla_w \mathcal{L}(w, b, \lambda) &= w - \sum_{i=1}^{N}\lambda_i y_i x_i = 0 \nonumber \\
    \therefore \quad w &= \sum_{i=1}^{N} \lambda_i y_i x_i \label{eq:grad-L-w}\\
    \frac{\partial \mathcal{L}(w,b,\lambda)}{\partial b} &= -\sum_{i=1}^{N} \lambda_i y_i = 0 \nonumber \\ 
    \therefore \quad \sum_{i=1}^{N} \lambda_i y_i &= 0 \label{eq:grad-L-b}
\end{align}

Substituting Eqs. \ref{eq:grad-L-w} and \ref{eq:grad-L-b} into Eq. \ref{eq:lagrangian-dual}, we have
\begin{align}
    \mathcal{L}(\lambda) = \sum_{i=1}^{N}\lambda_i - \frac{1}{2} \sum_{i=1}^{N}\sum_{j=1}^{N}\lambda_i \lambda_j x_i x_j y_i y_j \label{eq:lagrangian-dual-substituted}
\end{align}

Note that Eq. \ref{eq:lagrangian-dual-substituted} is a function of $\lambda$ only.
We want to maximize Eq. \ref{eq:lagrangian-dual-substituted} for the Lagrangian multipliers and ensure that $\lambda_i, \lambda_j \ge 0 \quad \forall i, j$ while satisfying Eq. \ref{eq:grad-L-b}.

\section{Formulation for AQCs}
\label{sub:svm-formulation}

AQCs are adept at solving QUBO problems, which are NP-hard and defined as follows:
\begin{align}
    \min_{z \in \mathbb{B}^M} z^T A z + z^T b, \label{eq:qubo}
\end{align}

where
$z \in \mathbb{B}^M$ is the binary decision vector ($M \in \mathbb{N}$),
$A \in \mathbb{R}^{M \times M}$ is the QUBO matrix, and
$b \in \mathbb{R}^M$ is the QUBO vector.

To convert the SVM training problem into a QUBO problem, we write Eq. \ref{eq:lagrangian-dual-substituted} as a minimization problem:
\begin{align}
    \min_{\lambda} \ \mathcal{L}(\lambda) &= \frac{1}{2} \sum_{i=1}^{N} \sum_{j=1}^{N} \lambda_i \lambda_j x_i x_j y_i y_j - \sum_{i=1}^{N} \lambda_i \label{eq:lagrangian-dual-min} \\
    \lambda_i, \lambda_j &\ge 0 \quad \forall i, j \nonumber 
\end{align}

This can be written in matrix form as follows:
\begin{align}
    \min_{\lambda} \mathcal{L}(\lambda) = \frac{1}{2} \lambda^T (X X^T \odot Y Y^T) \lambda - \lambda^T 1_N \qquad \lambda \ge 0_N, \label{eq:lagrangian-dual-matrix-form}
\end{align}

where $1_N$ and $0_N$ represent $N$-dimensional vectors of ones and zeros, respectively, and $\odot$ is the element-wise multiplication operation.

We now introduce a $K$-dimensional precision vector, $P = [p_1, p_2, \ldots, p_K]^T$, where each entry $p_k$ is a power of $2$.
This is required to impose the non-negativity constraint on $\lambda$.
The precision vector must also be sorted.
For example, a precision vector could be $P = \left[\frac{1}{4}, \frac{1}{2}, 1, 2, \right]^T$.
We also introduce $K$ binary variables $\hat{\lambda}_{ik}$ for each Lagrangian multiplier such that
\begin{align}
    % w_{j} &= \sum_{k=1}^K p_k \hat{w}_{jk} \qquad \forall j = 1, 2, \ldots, d \label{eq:binarized-svm-weights} \\
    % b &= \sum_{k=1}^K p_k \hat{b}_{k} \label{eq:binarized-svm-bias} \\
    \lambda_i &= \sum_{k=1}^K p_k \hat{\lambda}_{ik} \qquad \forall i = 1, 2, \ldots, N, \label{eq:binarized-svm-lagrange-multipliers}
\end{align}

where $p_k$ denotes the $k^{\text{th}}$ entry in the precision vector $P$.
% Summing from $K_+$ in Eq. \ref{eq:binarized-svm-lagrange-multipliers} ensures that the Lagrange multipliers are always positive, which is required when solving Problem \ref{eq:lagrangian-dual-min}.
% $\hat{w}_{jk}$, $\hat{b}_k$ and $\hat{\lambda}_{jk}$ can be thought of as binary decision variables that select or ignore entries in $P$ depending on whether their value is $1$ or $0$ respectively.
% With this formulation, we can have up to $2^K$ unique values for each $w_i$ when $P$ contains only positive values for instance.
% However, if $P$ contains negative values as well, then the number of unique attainable values for each $w_{i}$ might be less than $2^K$.
% For example, if $P = [-1, -\frac{1}{2}, \frac{1}{2}, 1]$, then only the following seven distinct values can be attained: $\{-\frac{3}{2}, -1, -\frac{1}{2}, 0, \frac{1}{2}, 1, \frac{3}{2}\}$.
Next, we vertically stack all binary variables:
\begin{align}
    % \hat{w} &= [\hat{w}_{11} \ \ldots \ \hat{w}_{1K} \ \hat{w}_{21} \ \ldots \ \hat{w}_{2K} \ \ldots \ \hat{w}_{d1} \ \ldots \ \hat{w}_{dK}]^T \label{eq:w-hat} \\
    % \hat{b} &= [\hat{b}_{1} \ \ldots \ \hat{b}_K]^T \label{eq:b-hat} \\
    \hat{\lambda} &= [\hat{\lambda}_{11} \ \ldots \ \hat{\lambda}_{1K} \ \hat{\lambda}_{21} \ \ldots \ \hat{\lambda}_{2K} \ \ldots \ \hat{\lambda}_{N1} \ \ldots \ \hat{\lambda}_{NK} ]^T \label{eq:lambda-hat}
\end{align}

We now define a precision matrix as follows:
\begin{align}
    % \mathcal{P} &=  \begin{bmatrix}
                    %     I_{d+1} \otimes P^T     & 0_{(d+1) \times N (K - K_+ + 1)} \\
                    %     0_{N \times (d + 1)}    & I_N \otimes P_+^T 
                    % \end{bmatrix} 
    \mathcal{P} = I_N \otimes P^T \label{eq:precision-matrix}
\end{align}

Notice that
\begin{align}
    \lambda &= \mathcal{P} \hat{\lambda} \label{eq:lambda-lambda-hat}
\end{align}

Finally, we substitute the value of $\lambda$ from Eq. \ref{eq:lambda-lambda-hat} into Eq. \ref{eq:lagrangian-dual-matrix-form}:
\begin{align}
    \min_{\hat{\lambda} \in \mathbb{B}^{NK}} \ \mathcal{L} (\hat{\lambda}) = \frac{1}{2} \hat{\lambda}^T \mathcal{P}^T (X X^T \odot Y Y^T) \mathcal{P} \hat{\lambda} - \hat{\lambda}^T \mathcal{P}^T 1_N \label{eq:svm-qubo}
    % \min_{\hat{\theta}} \ \mathcal{L} (\theta) &= \hat{\theta}^T \mathcal{P}^T U \mathcal{P} \hat{\theta} + \hat{\theta}^T  \mathcal{P}^T v \label{eq:svm-qubo}
\end{align}

Equation \ref{eq:svm-qubo} is identical to Eq. \ref{eq:qubo} with $z = \hat{\lambda}$, $A = \frac{1}{2}\mathcal{P}^T (X X^T \odot Y Y^T) \mathcal{P}$, $b = -\mathcal{P}^T 1_N$, and $M = KN$.
Hence, we converted the SVM training problem from Eq. \ref{eq:lagrangian-dual} into a QUBO problem in Eq. \ref{eq:svm-qubo}, which can then be solved on AQCs. 

\subsection{Computational Complexity}
\label{sub:svm-analysis}
We begin our theoretical analysis by defining the space complexity for the number of qubits needed to solve the QUBO. 
The SVM training problem stated in Eq. \ref{eq:lagrangian-dual-matrix-form} contains $\mathcal{O}(N)$ variables ($\lambda$) and $\mathcal{O}(Nd)$ data ($X$ and $Y$).
The QUBO formulation of the SVM training problem stated in Eq. \ref{eq:svm-qubo} consists of the same amount of data.
However, as part of the QUBO formulation, we introduced $K$ binary variables for each Lagrangian multiplier in the original problem (Eq. \ref{eq:lagrangian-dual-matrix-form}).
So, the total number of variables in Eq. \ref{eq:svm-qubo} is $\mathcal{O}(KN)$.
% The hardware connectivity of the D-Wave quantum annealer limits the size of QUBO problems that can be solved using the minimal number of hardware qubits.
% As a result, a direct 1-to-1 mapping of variables to hardware qubits is not possible.
% Using an efficient embedding algorithm such as \cite{date2019efficiently}, it is possible to embed a given QUBO problem over quadratic number of qubits. %(\fix{is there an upper bound on the size of the QUBO?})
So, the qubit footprint (or space complexity) of this formulation would be $\mathcal{O}(N^2 K^2)$ after embedding onto the hardware.
% In a practical setting, the number of data points is larger than the dimension of each data point, i.e. $N \gg d$. 
% Thus, the number of variables would be $\mathcal{O}(NK)$, and the qubit footprint would be $\mathcal{O}(N^2 K^2)$.

% \hl{
The time complexity of classical SVM algorithms is $\mathcal{O}(N^3)$ \cite{bottou2007support}.
% In order for us to convert the SVM training problem from Eq. \ref{eq:svm-qpp} to Eq. \ref{eq:svm-to-qubo}, we have to populate a square QUBO matrix of size equal to the number of variables ($\mathcal{O}(KN + Kd)$).
We analyze the time complexity for our approach in three parts.
First, the time complexity of converting Problem~\ref{eq:svm-qpp} into a QUBO problem can be inferred from Eqs. \ref{eq:lagrangian-dual-min} and \ref{eq:binarized-svm-lagrange-multipliers}
% }
% \begin{align}
%     \min_{\hat{w}, \hat{b}, \hat{\lambda}} \ \mathcal{L}(\hat{w}, \hat{b}, \hat{\lambda}) &= -\sum_{j=1}^d \sum_{k=1}^{K} \sum_{l=1}^{K} p_k p_l \hat{w}_{jk} \hat{w}_{jl} + \sum_{i=1}^{N} \sum_{j=1}^{d} \sum_{k=1}^{K} \sum_{l=K_+}^{K} x_{ij} y_i p_k p_l \hat{w}_{jk} \hat{\lambda}_{il} + \sum_{i=1}^{N} \sum_{k=1}^{K} \sum_{l=K_+}^{K} y_i p_k p_l \hat{b}_k \hat{\lambda}_{il}
%      - \sum_{i=1}^{N} \sum_{l=K_+}^K p_l \hat{\lambda}_{il} \label{eq:svm-qpp-summation}
% \end{align}
% % \hl{
% From Eq. \ref{eq:svm-qpp-summation}, the time complexity is 
as $\mathcal{O}(N^2K^2)$. 
% which is dominated by the second term.
% \hl{
Because we have $\mathcal{O}(NK)$ variables in the QUBO formulation, embedding can be done in $\mathcal{O}(N^2 K^2)$ time by using the embedding algorithm proposed by Date et al. \cite{date2019efficiently}.
Although the theoretical time complexity of quantum annealing used to obtain an exact solution is exponential ($\mathcal{O}(e^{\sqrt{d}})$) \cite{mukherjee2015multivariable}, a more realistic estimate of the running time can be made by using measures such as ST99 and ST99(OPT) \cite{wang2019simulated}, which give the expected number of iterations to reach a certain level of optimality with $99\%$ certainty.
Quantum annealing performs well on problems in which the energy barriers between local optima are tall and narrow because such an energy landscape is more conducive to quantum tunneling. 
To estimate ST99 and ST99(OPT) for our approach, details on specific instances of the SVM problem are required.
Estimating ST99 and ST99(OPT) for generic QUBO formulation of the SVM problem is beyond the scope of the present work.

That said, we would like to shed some light on the quantum annealing running times observed in practice.
% Obtaining the solution on AQCs depends on the annealing time, which is not $\mathcal{O}(1)$ in general, but can be treated as $\mathcal{O}(1)$ 
% A constant annealing time and a constant number of repetitions seem to work well for all practical purposes on an AQC of fixed and finite size, such as the D-Wave 2000Q \cite{date2019classical}. 
An AQC can accommodate only finite-sized problems---for example, D-Wave 2000Q can accommodate problems with 64 or fewer binary variables that require all-to-all connectivity \cite{date2019efficiently}.
For problems within this range, a constant annealing time and a constant number of repetitions seem to work well in practice.
So, the total time to convert and solve a linear regression problem on an AQC would be $\mathcal{O}(N^2 K^2)$.

% Secondly, the time taken to embed the $(NK)$-sized QUBO problem on the quantum computer is $\mathcal{O}(N^2K^2)$ (see Regression section for more details).
% Lastly, for the reasons mentioned in the Regression section, it is not straight forward to get a realistic estimate of the time complexity of the quantum annealing process.
% However, a constant annealing time in conjunction with a constant number of repetitions seems to work well in practice on an AQC of fixed and finite size as explained in the Regression section.
% So, the total time complexity is $\mathcal{O}(N^2K^2)$.

Note that the qubit footprint $\mathcal{O}(N^2K^2)$ and time complexity $\mathcal{O}(N^2K^2)$ assume that $K$ is a variable.
% However, the precision can be fixed universally, for example, $32$-bit or $64$-bit precision on classical computers.
% In such a scenario, $K$ can be treated as a constant. 
If the precision for all parameters ($\hat{\lambda}$) is fixed (e.g., limited to $32$-bit or $64$-bit precision), then $K$ becomes a constant factor.
The resulting qubit footprint would be $\mathcal{O}(N^2)$, and time complexity would also be be $\mathcal{O}(N^2)$.
This time complexity is an order of magnitude better than the classical algorithm ($\mathcal{O}(N^3)$).  
% }

\section{Empirical Analysis}
\label{sec:analysis}

\subsection{Methodology and Performance Metrics} 
Our investigation compares the accuracy of the classical SVM implemented in Scikit-learn with the quantum SVM that utilizes the D-Wave Advantage AQC. In addition to the quantum approach, we also consider simulated annealers to compare accuracy. Scikit-learn solves SVMs by using the sequential minimal optimization algorithm, whereas the quantum approach solves SVMs by transforming the problem into a QUBO problem. In our study, we convert Problem~\ref{eq:svm-qpp} to a Lagrangian dual (Problem~\ref{eq:lagrangian-dual-matrix-form}) and then into a QUBO (Problem~\ref{eq:svm-qubo}) for the quantum and simulated annealers. Our research evaluates the performance of the quantum approach, the classical approach, and the simulated annealing approach across two key metrics: (i) accuracy of the trained model and (ii) total compute time. We calculate the accuracy by dividing the number of correct classifications by the total number of samples. For the classical approach, we measure the compute time as the time taken for training. In contrast, the compute time for the quantum approach encompasses the time required for converting the problem into a QUBO problem, embedding the QUBO problem onto the quantum hardware, and performing quantum annealing to solve the problem. Notably, we observed that the simulated annealers failed to find a viable solution, especially when there were many features,  
so we decided not to examine the computing time scalability for simulated annealers.

\subsection{Hardware Configuration}
We conducted the implementations of the classical approach, simulated annealing, and preprocessing stage on two different hardware configurations without altering the underlying methodology.
The first system, equipped with an AMD Ryzen 4600H 6-core CPU running at 3~GHz and 16~GB of DDR4 RAM running at 2,666 MHz, was used to evaluate the accuracy results described in Section~\ref{sub:accuracy-results}. 
The second system, equipped with an Intel Xeon E5-2690v4 14-core CPU running at 2.6~GHz and 64~GB of DDR4 RAM running at 2,400 MHz, was used to evaluate the compute times presented in Section~\ref{sub:scalability-results}. 
We used the two different machines for the accuracy and compute time comparisons because the AMD machine could not support the larger synthetic datasets used in the compute time experiments.
% The latter machine was we compare time performance on a separate configuration from the accuracy. 

We used the D-Wave Advantage quantum annealer to evaluate the quantum approach; the annealer offers 5,627 functioning qubits. After preprocessing the problem on a classical computer, we employed the \texttt{EmbeddingComposite} class from the D-Wave library to embed the problem onto the quantum hardware. We conducted 10 annealing runs for each problem instance to ensure accuracy and selected the solution with the lowest energy sample.

\subsection{Datasets}
We compare the classical, quantum, and simulated annealing approaches across the following datasets: synthetic datasets, Iris, WBC, Wine, Digits, and Lambeq. When datasets were split for training and testing, the training data was uniformly split per class.
For generating the synthetic datasets, we employ two methods.
The first method utilizes the \texttt{make\_blob} function from the Scikit-learn library. 
This function generates synthetic data with two linearly separable centers and $d$-dimensional sides. In the second method, we implement a function that generates data points and their corresponding class labels based on the following specified parameters: number of data points, number of features, SVM weights, and SVM bias. 
We generate each data point by sampling the feature values uniformly at random in the range [$-$1, 1]. We then classify these points based on a linear decision boundary. 
This process iteratively generates the data until we have the desired number of data points. 
This approach allows us to generate synthetic datasets that, when trained with an SVM, result in predictable hyperplanes.
The Iris dataset contains 150 data points with four features, and the WBC dataset consists of 369 data points with 30 features. 
The Wine dataset contains 178 data points with 13 features for three classes, and the Digits dataset holds 1,797 data points with 64 features and 10 classes, but we only used two classes of 168 data points. 

\begin{figure}
    \centering
    \includegraphics[width=\linewidth]{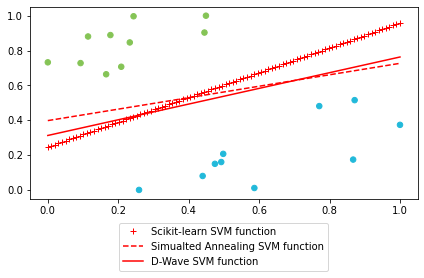}
    \caption{Comparison of hyperplanes created by the support vectors with Scikit-learn (+), simulated annealing(- -), and D-Wave (---) on positive synthetic data (blue and green circles).}
    \label{fig:syn_pos}
\end{figure}
\begin{figure}
    \centering
    \includegraphics[width=\linewidth]{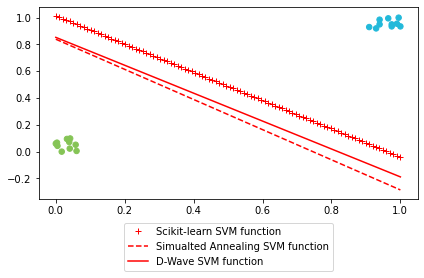}
    \caption{Comparison of hyperplanes created by the support vectors with Scikit-learn (+), simulated annealing(- -) and D-Wave (---) on negative synthetic data (blue and green circles).}
    \label{fig:syn_neg}
\end{figure}
\begin{figure}
    \centering
    \includegraphics[width=\linewidth]{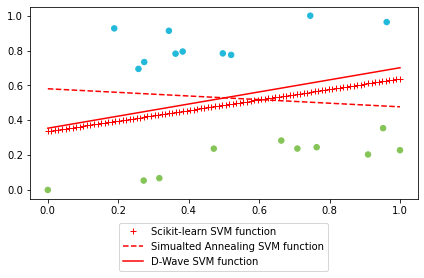}
    \caption{Comparison of hyperplanes created by the support vectors with Scikit-learn (+), simulated annealing(- -), and D-Wave (---) on random synthetic data (blue and green circles split by classification).}
    \label{fig:syn_var4}
\end{figure}

\begin{table*}[t!]
\caption{Training and Testing Accuracy of Datasets}
\label{tab:accuracy}
 \setlength{\tabcolsep}{4pt}
  \scriptsize
\begin{tabular}{lllllllll}
\toprule
Data &
  Trained &
  Tested &
  Scikit-learn &
  Scikit-learn &
  D-Wave &
  D-Wave  &
  Simulated Annealing  &
  Simulated Annealing \\
  & Points & Points & Training Accuracy & Testing Accuracy
  & Training Accuracy & Testing Accuracy & Training Accuracy & Testing Accuracy \\
  \midrule
Synthetic positive              & 20 & 100 & 100    & 100     & 100     & 100     & 100     & 99.6 ± 0.8 \\
Synthetic negative              & 20 & 100 & 100     & 100     & 100     & 100     & 100     & 100     \\
Synthetic random              & 20 & 100 & 100     & 100     & 100     & 100     & 100     & 100     \\
Setosa - Virginica     & 20 & 80  & 100     & 100     & 100    & 100     & 100     & 99.6 ± 1.2 \\
Setosa - Versicolor    & 20 & 80  & 100     & 99.6 ± 0.6 & 100     & 99.5 ± 1.6 & 100     & 99.7 ± 0.5 \\
Versicolor - Virginica & 20 & 80  & 90.5 ± 6  & 85.6 ± 3.3 & 70.5 ± 33.1 & 67.9 ± 28.8 & 93.5 ± 6.7 & 88.1 ± 6.7 \\
WBC                   & 52 & 517 & 97.7 ± 1.2 & 95 ± 1.4  & 93.3 ± 21 & 93.1 ± 1.3 & 91.7 ± 2.6 & 92.6 ± 1.4 \\
Wine 0|1                 & 52 & 55  & 100     & 100    & 100  & 100   & 99 ± 1.4  & 99.5 ± 1.7 \\
Wine 0|2                 & 52 & 78  & 100     & 98.2 ± 1.8 & 96 ± 2.1  & 96 ± 1.3  & 95 ± 2.1  & 94.7 ± 2.3 \\
Wine 1|2                 & 52 & 67  & 99.2 ± 1  & 96 ± 1.7  & 98.5 ± 1.5 & 96.6 ± 2.4 & 96 ± 4   & 94.2 ± 4.5\\
Digits 0|1              & 20 & 158 & 100    & 99.2 ± 1.3 & 99 ± 2.1  & 97.5 ± 2.2 & --    & --   \\
Lambeq                  & 52 & 30  & 100  & 100  & 93.5  ± 2.9 & 88.7 ± 4.8 & --   & -- \\
\bottomrule
\end{tabular}
\end{table*}

\subsection{Accuracy}
\label{sub:accuracy-results}
We compare the classification accuracy of three distinct SVM approaches: the classical SVM, simulated annealing SVM, and quantum SVM. We employ multiple datasets and conduct 10 rounds of training and testing for each dataset.
In many of these datasets, we use only 52 points chosen uniformly at random for training the classifiers because 52 is the largest size of the training dataset that could be accommodated on the annealer.
The accuracy results are presented in Table \ref{tab:accuracy}.

\subsubsection{Synthetic Data}
We parameterize the synthetic datasets into three types: positive, negative, and random. Positive datasets are designed to be linearly separable when separated by a positive hyperplane (see Figure \ref{fig:syn_pos}). Conversely, negative datasets are designed to be separed by a negative hyperplane (see Figure \ref{fig:syn_neg}). To achieve this, we classify randomly sampled points based on a linear decision boundary. Additionally, random datasets with linearly separable data points are generated using the Scikit-learn \texttt{make\_blob} function (see Figure \ref{fig:syn_var4}). Before training the SVM models, we ensure that all data points are normalized. Each training dataset consistently contains 20 points with two features. For testing accuracies, we use 100 points. We provide a detailed comparison of scaling data points and features later in this work.

All three approaches---classical SVM, simulated annealing SVM, and quantum SVM---accurately identify the support vectors for the synthetic datasets. Figures~\ref{fig:syn_pos}, \ref{fig:syn_neg}, and \ref{fig:syn_var4} illustrate the datasets and corresponding hyperplanes constructed by the support vectors. Both classical SVM and quantum SVM exhibit $100\%$ accuracy during training and testing across all three datasets. The simulated annealing approach also achieved \(100\%\) accuracy on the negative and random datasets. For the positive datasets, the simulated annealing approach achieved an average accuracy of \(99.6\%\). These results highlight the effectiveness of all three methods in accurately identifying support vectors. 

\begin{table*}[t!]
\caption{\textbf{Training and Testing Accuracy of Varying Annealing Time}}
\label{tab:annealing_time}
 \setlength{\tabcolsep}{4pt}
  \scriptsize
\begin{tabular}{llllllllll}
\toprule
Data &
  Annealing&
  Trained &
  Tested &
  Scikit-learn &
  Scikit-learn &
  D-Wave &
  D-Wave  &
  Simulated Annealing  &
  Simulated Annealing \\
  & Time(µs) & Points & Points & Training Accuracy & Testing Accuracy
  & Training Accuracy & Testing Accuracy & Training Accuracy & Testing Accuracy \\
  \midrule
Versicolor - Virginica&20&52&48&93.8 ± 3.1&94.2 ± 3.2&87.5 ± 5.4&86.2 ± 6.4&87.5 ± 4.6&85.4 ± 4.7\\
Versicolor - Virginica&100&52&48&95.0 ± 2.3&93.1 ± 4.3&88.1 ± 4.1&85.8 ± 5.8&87.1 ± 3.4&86.7 ± 5.0\\
Versicolor - Virginica&1000&52&48&93.7 ± 2.4&95.0 ± 2.2&86.7 ± 2.8&89.8 ± 4.0&88.1 ± 3.2&89.0 ± 3.9\\
WBC&20&52&517&95.6 ± 2.0&95.7 ± 0.9&90.8 ± 3.9&92.7 ± 1.2&91.0 ± 3.4&93.2 ± 0.7\\
WBC&100&52&517&97.3 ± 2.6&95.4 ± 1.3&90.6 ± 3.2&92.9 ± 1.1&91.3 ± 2.8&93.2 ± 1.0\\
WBC&1000&52&517&96.7 ± 1.3&95.4 ± 1.1&91.3 ± 3.9&92.2 ± 1.1&91.7 ± 3.1&92.4 ± 1.7\\
Wine 1|2&20&52&67&97.9 ± 1.9&97.2 ± 2.0&97.3 ± 2.7&97.0 ± 1.4&95.0 ± 4.3&96.3 ± 3.3\\
Wine 1|2&100&52&67&98.7 ± 1.8&96.3 ± 1.8&97.5 ± 2.6&96.0 ± 2.9&96.7 ± 3.6&96.7 ± 2.2\\
Wine 1|2&1000&52&67&98.8 ± 1.3&97.2 ± 2.2&96.7 ± 2.0&95.7 ± 2.9&97.1 ± 1.6&95.7 ± 4.3\\

\bottomrule
\end{tabular}
\end{table*}
\subsubsection{Iris}
The Iris dataset consists of 150 samples, three classes, and four features. The three classes describe the Setosa, Versicolor, and Virginica flowers, and the four features represent the length and width of the petal and sepal of the flower. The objective is to perform binary classification for each pair of classes: (Setosa, Versicolor), (Setosa, Virginica), and (Versicolor, Virginica). We normalize the data for training. For each pair, we utilize 20 data points across both classes for training and evaluate the model's accuracy by using the remaining 80 points.
The accuracy results are shown in Table \ref{tab:accuracy}.

First, all three approaches classify the Setosa and Virginica flowers well. The classical and quantum approaches achieve \(100\%\) accuracy in the training and testing data classification. The simulated annealer achieves \(100\%\) accuracy in the training data and \(99.6\%\) accuracy in the testing data.
Second, all three approaches demonstrate near perfect classification performance for Setosa and Versicolor flowers. 
The classical approach attains an average of \(100\%\) training accuracy and \(99.6\%\) testing accuracy. The quantum annealer achieves \(100\%\) training accuracy and \(99.5\%\) testing accuracy. The simulated annealer attains \(100\%\) training accuracy and \(99.7\%\) testing accuracy.  Notably, there is no statistically significant difference in accuracy across the three training methods.
Third, all three methods exhibit lower classification performance for the Versicolor and Virginica flower species, which are known to be linearly inseparable.
The classical approach achieves \(90.5\%\) training accuracy and \(85.6\%\) testing accuracy. The quantum annealer achieves \(70.5\%\) training accuracy and \(67.9\%\) testing accuracy. The simulated annealer achieves \(93.5\%\) training accuracy and \(88.1\%\) testing accuracy. 

\subsubsection{WBC}
The WBC dataset consists of 569 samples, two classes, and 30 features. The data represents digitalized characteristics of breast cell nuclei. The data is normalized for training. For evaluation, 52 points are used for training, and 517 points are used for testing accuracy.
The classical approach demonstrates the highest accuracy and achieves an average of \(97.7\%\) for training data and \(95.0\%\) for testing data. The quantum approach follows with a training accuracy of \(93.3\%\) and a testing accuracy of \(93.1\%\). Last, the simulated annealing approach achieves a training accuracy of \(91.7\%\) and a testing accuracy of \(92.6\%\).

\subsubsection{Wine}
The Wine dataset has 178 samples, three classes, and 13 features. The features describe characteristics of the wine, including alcohol, ash, and magnesium content. We compare the three classes in combination: (class 0 and class 1), (class 0 and class 2), and (class 1 and class 2). In Table~\ref{tab:accuracy}, the classes are represented as Wine Class \# - Class \#. The data is normalized for training, and 52 points are used. The remaining points are used for testing accuracy for the three combinations of datasets.

First, when comparing class 0 and class 1, 52 points are trained and 55 points are tested. The training and testing accuracy averaged \(100\%\) for the classical SVM and quantum SVM and  \(99\%\) and \(99.5\%\) for the simulated annealer. 
Second, when comparing class 0 and class 2, 52 points are trained and 78 points are tested. The training and testing accuracy averaged \(100\%\) and \(98.2\%\) for the classical SVM. The quantum SVM averaged \(96\%\) in accuracy for both training and training, and the simulated annealing SVM averaged \(95.0\%\) and \(94.7\%\) in accuracy.
Third, when comparing class 1 and class 2, 52 points are trained and 67 points are tested. The training and testing accuracy averaged \(99.2\%\) and \(96.0\%\) for the classical SVM,  \(98.5\%\) and \(96.6\%\) for the quantum SVM, and  \(96.0\%\) and \(94.2\%\) for the simulated annealer.

\begin{table*}[t!]
  \centering
  \caption{Scalability with Number of Features}
  \label{tab:features}
  \setlength{\tabcolsep}{4pt}
  \scriptsize
  \begin{tabular}{lllllll}
    \toprule
    
    \textbf{Features} & \textbf{Scikit-learn} & \textbf{D-Wave Embedding} & \textbf{D-Wave Preprocessing} & \textbf{D-Wave Access} & \textbf{D-Wave Compute} & \textbf{D-Wave Network}  \\
     & \textbf{Time (ms)} & \textbf{Time (ms)} & \textbf{Time (ms)} & \textbf{Time (ms)} & \textbf{Time (ms)} & \textbf{Overhead (ms)} \\
    & & & & & \textbf{(Preprocess +} \\
    & & & & & \textbf{Embedding + Access)}  \\
    \midrule
    2 & \textbf{1.175 $\pm$ 0.072} & 1,379.469 $\pm$ 203.994 & 239.732 $\pm$ 28.529 & 18.39 $\pm$ 0.057 & \textbf{1,637.592 $\pm$ 211.0} & 422,713.522 $\pm$ 111,923.977 \\
    4 & \textbf{1.158 $\pm$ 0.024 }& 1,405.793 $\pm$ 377.299 & 231.087 $\pm$ 1.612 & 18.33 $\pm$ 0.095 &\textbf{1,655.21 $\pm$ 377.127} & 330,455.173 $\pm$ 90,493.002 \\
    8 &\textbf{1.185 $\pm$ 0.047} & 1,164.539 $\pm$ 378.807 & 231.308 $\pm$ 1.815 & 18.29 $\pm$ 0.099 & \textbf{1,414.137 $\pm$ 378.851} & 450,387.29 $\pm$ 206,827.975\\
    16 &\textbf{1.192 $\pm$ 0.014} & 1,232.807 $\pm$ 568.895 & 232.896 $\pm$ 1.886 & 18.37 $\pm$ 0.048 &\textbf{1,484.073 $\pm$ 568.916} & 322,623.635 $\pm$ 106,189.784 \\
    32 &\textbf{1.18 $\pm$ 0.09} & 1,120.983 $\pm$ 97.316 & 231.457 $\pm$ 1.17 & 18.39 $\pm$ 0.057 & \textbf{1,370.829 $\pm$ 97.337} & 370,004.535 $\pm$ 126,229.111 \\
    64 &\textbf{1.253 $\pm$ 0.098} & 1,381.844 $\pm$ 839.621 & 243.054 $\pm$ 30.388 & 18.38 $\pm$ 0.042 &\textbf{1,643.278 $\pm$ 836.636} & 552,217.955 $\pm$ 286,747.464  \\
    128 & \textbf{1.249 $\pm$ 0.043 }& 1,073.689 $\pm$ 54.481 & 234.919 $\pm$ 11.097 & 18.33 $\pm$ 0.067 & \textbf{1,326.939 $\pm$ 59.699 }& 418,365.701 $\pm$ 239,712.534\\
    256 &\textbf{1.206 $\pm$ 0.082 }& 1,386.806 $\pm$ 161.495 & 234.472 $\pm$ 29.823 & 18.37 $\pm$ 0.048 & \textbf{1,639.648 $\pm$ 164.609} & 310,426.496 $\pm$ 138,119.185 \\
    512 &\textbf{1.261 $\pm$ 0.031} & 1,502.025 $\pm$ 388.979 & 230.431 $\pm$ 2.023 & 18.35 $\pm$ 0.071 &\textbf{1,750.806 $\pm$ 388.302} & 317,514.671 $\pm$ 156,945.844\\
    1,024 & \textbf{1.493 $\pm$ 0.075} & 1,423.757 $\pm$ 355.441 & 236.499 $\pm$ 16.529 & 18.29 $\pm$ 0.11 & \textbf{1,678.546 $\pm$ 353.131} & 390,337.469 $\pm$ 123,918.024  \\
    2,048 & \textbf{2.011 $\pm$ 0.145} & 1,646.613 $\pm$ 727.704 & 258.914 $\pm$ 42.713 & 18.34 $\pm$ 0.07 & \textbf{1,923.867 $\pm$ 720.138 }& 288,349.625 $\pm$ 135,981.113 \\
    4,096 & \textbf{4.598 $\pm$ 0.704 }& 1,482.837 $\pm$ 686.698 & 257.999 $\pm$ 40.914 & 18.34 $\pm$ 0.097 &\textbf{1,759.176 $\pm$ 681.247} & 399,891.987 $\pm$ 196,877.404 \\
    8,192 &\textbf{5.228 $\pm$ 0.888} & 1,280.656 $\pm$ 72.575 & 260.521 $\pm$ 40.755 & 18.39 $\pm$ 0.074 &\textbf{1,559.567 $\pm$ 105.107 }& 316,780.627 $\pm$ 139,171.499  \\
    16,384 &\textbf{9.234 $\pm$ 0.907 }& 1,321.959 $\pm$ 90.673 & 249.479 $\pm$ 35.551 & 18.37 $\pm$ 0.048 &\textbf{1,589.807 $\pm$ 112.371 }& 393,245.32 $\pm$ 97,614.288  \\
    32,768 & \textbf{16.969 $\pm$ 1.247} & 1,577.783 $\pm$ 1013.991 & 242.255 $\pm$ 23.858 & 18.36 $\pm$ 0.107 &\textbf{1,838.398 $\pm$ 1007.663 }& 347,998.856 $\pm$ 129,216.235 \\
    65,536 & \textbf{34.06 $\pm$ 2.547} & 1,692.453 $\pm$ 1249.103 & 234.506 $\pm$ 3.644 & 18.37 $\pm$ 0.048 &\textbf{1,945.328 $\pm$ 1248.834} & 387,732.537 $\pm$ 135,155.523 \\
    131,072 & \textbf{76.238 $\pm$ 4.305} & 1,302.487 $\pm$ 84.587 & 254.051 $\pm$ 44.079 & 18.34 $\pm$ 0.07 &\textbf{1,574.878 $\pm$ 93.238} & 303,776.797 $\pm$ 82,509.468 \\
    262,144 & \textbf{182.478 $\pm$ 10.178 }& 1,654.28 $\pm$ 1011.998 & 243.556 $\pm$ 31.683 & 18.37 $\pm$ 0.062 & \textbf{1,916.393 $\pm$ 1013.422 }& 356,133.736 $\pm$ 129,222.717  \\
    524,288 & \textbf{430.502 $\pm$ 26.294} & 1,530.793 $\pm$ 910.643 & 253.15 $\pm$ 42.777 & 18.37 $\pm$ 0.058 &\textbf{1,791.273 $\pm$ 908.992 }& 334,643.888 $\pm$ 109,913.759  \\
    1,048,576 & \textbf{822.135 $\pm$ 23.357} & 1,389.431 $\pm$ 115.631 & 281.621 $\pm$ 39.667 & 18.36 $\pm$ 0.07 & \textbf{1,689.412 $\pm$ 132.526} & 404,630.828 $\pm$ 274,147.286 \\
    2,097,152 &\textbf{1,543.618 $\pm$ 139.181} & 1,318.885 $\pm$ 88.77 & 294.537 $\pm$ 42.083 & 18.35 $\pm$ 0.071 & \textbf{1,631.772 $\pm$ 107.334 }& 328,081.545 $\pm$ 142,239.908 \\
    4,194,304 & \textbf{3,166.03 $\pm$ 316.12} & 1,300.346 $\pm$ 128.42 & 294.486 $\pm$ 32.706 & 18.36 $\pm$ 0.084 & \textbf{1,613.192 $\pm$ 124.671 }& 420,615.812 $\pm$ 195,816.586 \\
    8,388,608 & \textbf{6,920.407 $\pm$ 329.305} & 1,516.167 $\pm$ 332.535 & 343.214 $\pm$ 19.5 & 18.36 $\pm$ 0.052 & \textbf{1,877.741 $\pm$ 326.011} & 428,409.075 $\pm$ 256,341.033 \\

    \bottomrule
  \end{tabular}
\end{table*}

\begin{figure*}[t!]
    \centering
    \begin{subfigure}{\textwidth}
        \centering
        \includegraphics[width=\textwidth]{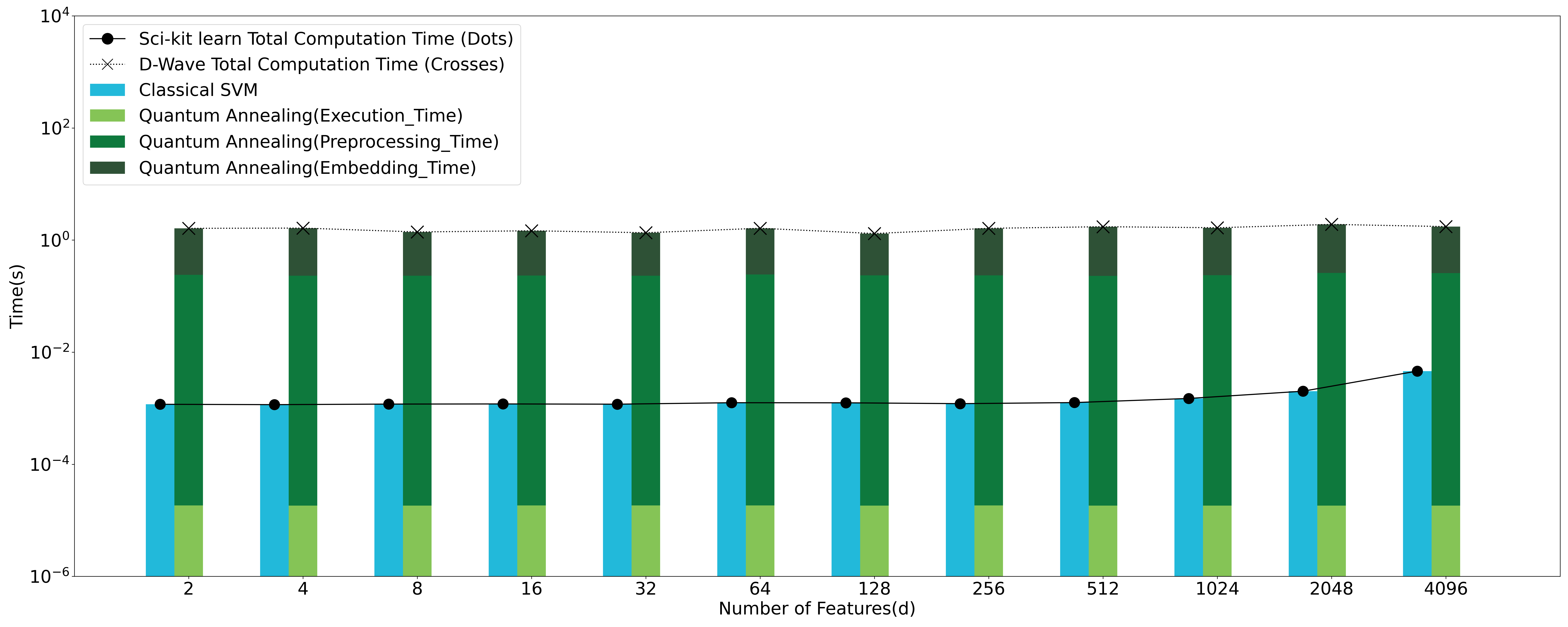}
        \caption{Few features ($d$)}
        \label{fig:svm-scaling-d-small}
    \end{subfigure}
    \begin{subfigure}{\textwidth}
        \centering
        \includegraphics[width=\textwidth]{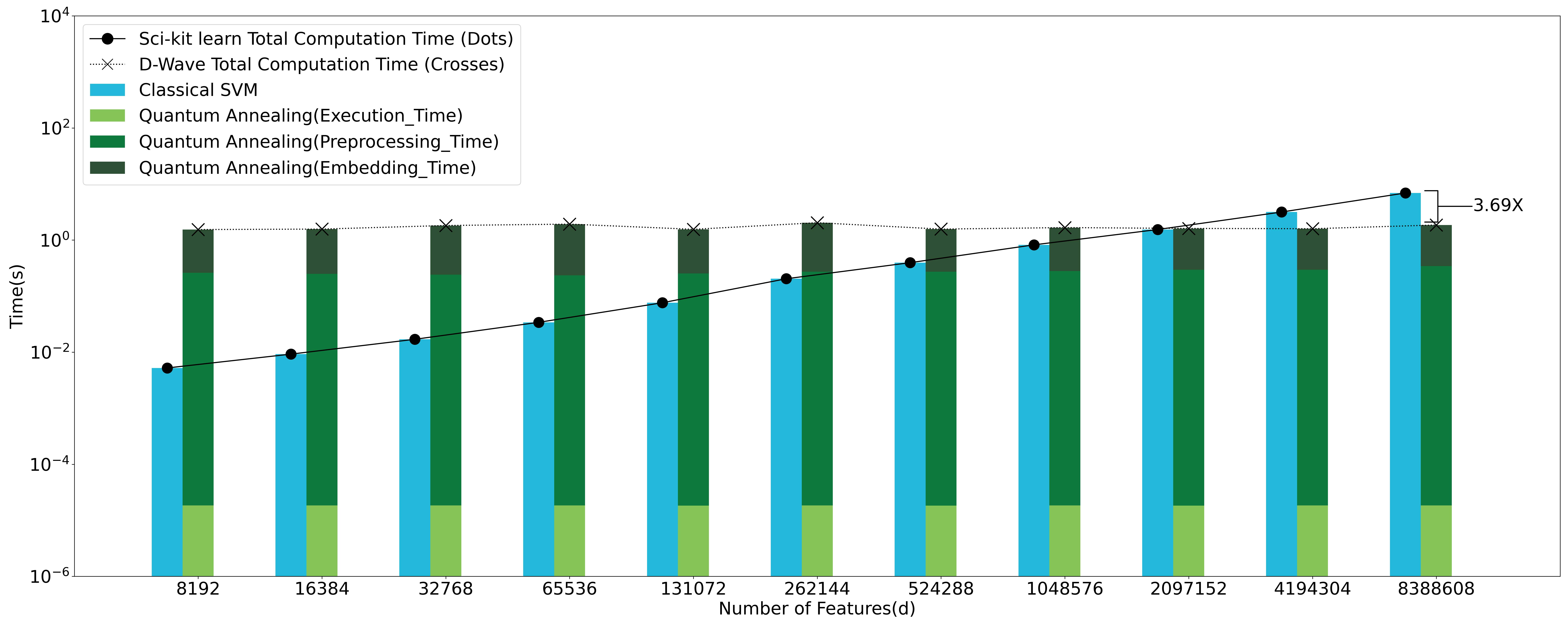}
        \caption{Many features ($d$)}
        \label{fig:svm-scaling-d-large}
    \end{subfigure}
    \caption{Scalability comparison of the Scikit-learn SVM (blue bar and bold line) and quantum SVM (light, medium, and dark green bars and dotted line). The $x$-axis indicates the number of features ($d$), and the $y$-axis logarithmically represents time in seconds. The $x$-axis ranges from $2$ to $8,388,608$ across the two figures. In Figure \ref{fig:svm-scaling-d-small}, $d$ varies between $2$ and $4,096$, and in Figure \ref{fig:svm-scaling-d-large}, $d$ varies between $8,192$ and $8,388,608$. In Figure \ref{fig:svm-scaling-d-large}, at 8 million features, the quantum approach demonstrated a speedup of 3.69$\times$ over the classical approach.}
    \label{fig:svm-scaling-d}
\end{figure*}

\begin{table*}[t!]
\centering
\caption{Scalability with Number of Points}
\label{tab:trained-points}
\scriptsize
\setlength{\tabcolsep}{4pt}
\begin{tabular}{lllllll}
\toprule
\textbf{Trained Points} & \textbf{Scikit-learn} & \textbf{D-Wave Embedding} & \textbf{D-Wave Preprocessing} & \textbf{D-Wave Access} & \textbf{D-Wave Compute} & \textbf{D-Wave Network}  \\
     & \textbf{Time (ms)} & \textbf{Time (ms)} & \textbf{Time (ms)} & \textbf{Time (ms)} & \textbf{Time (ms)} & \textbf{Overhead (ms)} \\
    & & & & & \textbf{(Preprocess +} \\
    & & & & & \textbf{Embedding + Access)}  \\
\midrule
4 & \textbf{397.597 $\pm$ 12.209} & 1,430.915  $\pm$  297.767 & 52.004 $\pm$ 7.233 & 17.0 $\pm$ 0.156 &\textbf{1,499.919  $\pm$ 297.346} & 120.693 $\pm$ 15.664 \\
6 & \textbf{628.511 $\pm$ 7.604} & 1,422.757 $\pm$ 382.261 & 71.3 $\pm$ 1.359 & 16.92 $\pm$ 0.187 & \textbf{1,510.977 $\pm$ 382.923} & 440.845 $\pm$ 81.854\\
8 & \textbf{865.129 $\pm$ 14.27} & 1,303.998 $\pm$ 516.343 & 83.676 $\pm$ 9.532 & 16.95 $\pm$ 0.217 &\textbf{1,404.623 $\pm$ 515.02} & 1,112.126 $\pm$ 360.746 \\
10 & \textbf{1,131.034 $\pm$ 12.523} & 1,633.786  $\pm$ 985.969 & 122.318 $\pm$ 14.208 & 17.01 $\pm$ 0.311 &\textbf{1,773.115 $\pm$ 992.71} & 2,049.21 $\pm$ 661.929 \\
12 & \textbf{1,414.692 $\pm$ 42.061} & 1,334.734 $\pm$ 177.382 & 141.291 $\pm$ 11.683 & 17.03 $\pm$ 0.134 &\textbf{1,493.055 $\pm$ 179.878} & 3,905.956 $\pm$ 1,651.884 \\
14 & \textbf{1,709.829 $\pm$ 30.631 }& 1,378.001 $\pm$ 380.399 & 168.314 $\pm$ 2.797 & 17.29 $\pm$ 0.179 & \textbf{1,563.604 $\pm$ 382.467} & 8,745.695 $\pm$ 3,737.998 \\
16 & \textbf{2,057.462 $\pm$ 21.925} & 1,154.179 $\pm$ 311.386 & 49.982 $\pm$ 2.321 & 17.18 $\pm$ 0.244 &\textbf{1,221.341 $\pm$ 310.555} & 10,334.816 $\pm$ 3,908.58 \\
18 & \textbf{2,298.212 $\pm$ 102.983} & 1,217.663 $\pm$ 419.797 & 66.299 $\pm$ 11.212 & 17.1 $\pm$ 0.189 &\textbf{1,301.062 $\pm$ 419.291 }& 19,508.428 $\pm$ 7,737.284 \\
20 & \textbf{2,623.09 $\pm$ 145.095} & 1,257.566 $\pm$ 530.37 & 72.883 $\pm$ 11.114 & 17.32 $\pm$ 0.266 & \textbf{1,347.769 $\pm$ 527.727 }& 23,161.555 $\pm$ 11,342.062 \\
22 & \textbf{3,117.991 $\pm$ 116.612} & 1,370.357 $\pm$ 674.162 & 84.43 $\pm$ 13.221 & 17.47 $\pm$ 0.291 &\textbf{1,472.257 $\pm$ 686.392 }& 38,135.434 $\pm$ 20,395.38 \\
24 & \textbf{3,402.623 $\pm$ 140.536} & 1,225.826 $\pm$ 309.729 & 91.573 $\pm$ 4.39 & 17.62 $\pm$ 0.215 & \textbf{1,335.019 $\pm$ 308.578 }& 39,167.391 $\pm$ 25,129.559 \\
26 & \textbf{3,854.724 $\pm$ 161.398 }& 1,078.185 $\pm$ 58.742 & 102.241 $\pm$ 3.995 & 17.66 $\pm$ 0.143 & \textbf{1,198.086 $\pm$ 59.46} & 49,917.087 $\pm$ 24,788.842 \\
28 & \textbf{4,158.047 $\pm$ 237.876 }& 1,409.069 $\pm$ 1,002.37 & 116.227 $\pm$ 2.926 & 17.54 $\pm$ 0.272 & \textbf{1,542.835 $\pm$ 1,003.154} & 74,291.927 $\pm$ 38,792.915 \\
30 & \textbf{4,325.487 $\pm$ 132.322} & 1,442.387 $\pm$ 1,252.714 & 145.596 $\pm$ 30.139 & 17.85 $\pm$ 0.085 & \textbf{1,605.834 $\pm$ 1,247.755} & 84,428.49 $\pm$ 44,035.425 \\
32 & \textbf{4,736.955 $\pm$ 146.286} & 1,074.25 $\pm$ 54.508 & 150.959 $\pm$ 23.223 & 17.87 $\pm$ 0.048 &\textbf{1,243.08 $\pm$ 63.392} & 109,971.795 $\pm$ 33,452.821 \\
34 & \textbf{4,895.237 $\pm$ 209.984} & 1,099.025 $\pm$ 75.892 & 173.559 $\pm$ 28.753 & 17.97 $\pm$ 0.095 & \textbf{1,290.554 $\pm$ 65.265} & 100,535.682 $\pm$ 52,262.449 \\
36 & \textbf{5,259.747 $\pm$ 252.264} & 2,015.937 $\pm$ 2,794.639 & 183.582 $\pm$ 23.139 & 17.93 $\pm$ 0.125 &\textbf{2,217.45 $\pm$ 2,791.223} & 112,656.481 $\pm$ 54,607.342 \\
38 & \textbf{5,390.033 $\pm$ 195.683} & 1,079.114 $\pm$ 59.131 & 217.531 $\pm$ 35.153 & 17.98 $\pm$ 0.169 & \textbf{1,314.624 $\pm$ 62.542} & 181,915.305 $\pm$ 142,277.523 \\
40 & \textbf{5,656.504 $\pm$ 101.086} & 1,868.626 $\pm$ 2,396.907 & 214.248 $\pm$ 7.722 & 18.1 $\pm$ 0.067 & \textbf{2,100.973 $\pm$ 2,396.668} & 215,169.1 $\pm$ 130,688.927 \\
42 & \textbf{5,962.922 $\pm$ 125.236} & 1,156.912 $\pm$ 102.021 & 237.135 $\pm$ 11.716 & 18.1 $\pm$ 0.105 & \textbf{1,412.147 $\pm$ 101.156} & 175,136.911 $\pm$ 66,927.252 \\
44 & \textbf{6,253.6 $\pm$ 91.927 }& 1,142.343 $\pm$ 87.118 & 246.837 $\pm$ 7.396 & 18.16 $\pm$ 0.084 & \textbf{1,407.339 $\pm$ 91.611} & 279,223.122 $\pm$ 119,973.036 \\
46 & \textbf{6,463.876 $\pm$ 145.813} & 1,179.851 $\pm$ 110.547 & 291.388 $\pm$ 29.706 & 18.18 $\pm$ 0.092 & \textbf{1,489.419 $\pm$ 106.889} & 248,253.607 $\pm$ 83,297.36 \\
48 & \textbf{6,771.785 $\pm$ 130.662} & 2,191.238 $\pm$ 3,049.933 & 292.981 $\pm$ 12.052 & 18.31 $\pm$ 0.074 & \textbf{2,502.53 $\pm$ 3,046.364} & 297,207.318 $\pm$ 178,830.42 \\
50 & \textbf{7,080.465 $\pm$ 96.879} & 1,222.904 $\pm$ 114.113 & 328.075 $\pm$ 36.607 & 18.3 $\pm$ 0.125 &\textbf{1,569.279 $\pm$ 104.291} & 292,123.268 $\pm$ 141,273.153 \\
52 & \textbf{7,194.556 $\pm$ 195.74} & 1,219.074 $\pm$ 88.527 & 369.37 $\pm$ 39.163 & 18.34 $\pm$ 0.07 &\textbf{1,606.784 $\pm$ 90.774} & 407,913.426 $\pm$ 209,009.574 \\
54 & \textbf{7,623.851 $\pm$ 530.636} & 4,031.514 $\pm$ 8,823.785 & 383.31 $\pm$ 34.857 & 18.38 $\pm$ 0.079 &\textbf{4,433.204 $\pm$ 8,824.813} & 399,734.428 $\pm$ 191,104.466 \\
\bottomrule
\end{tabular}
\end{table*}

\begin{figure*}[t!]
  \centering
  \includegraphics[width=\textwidth]{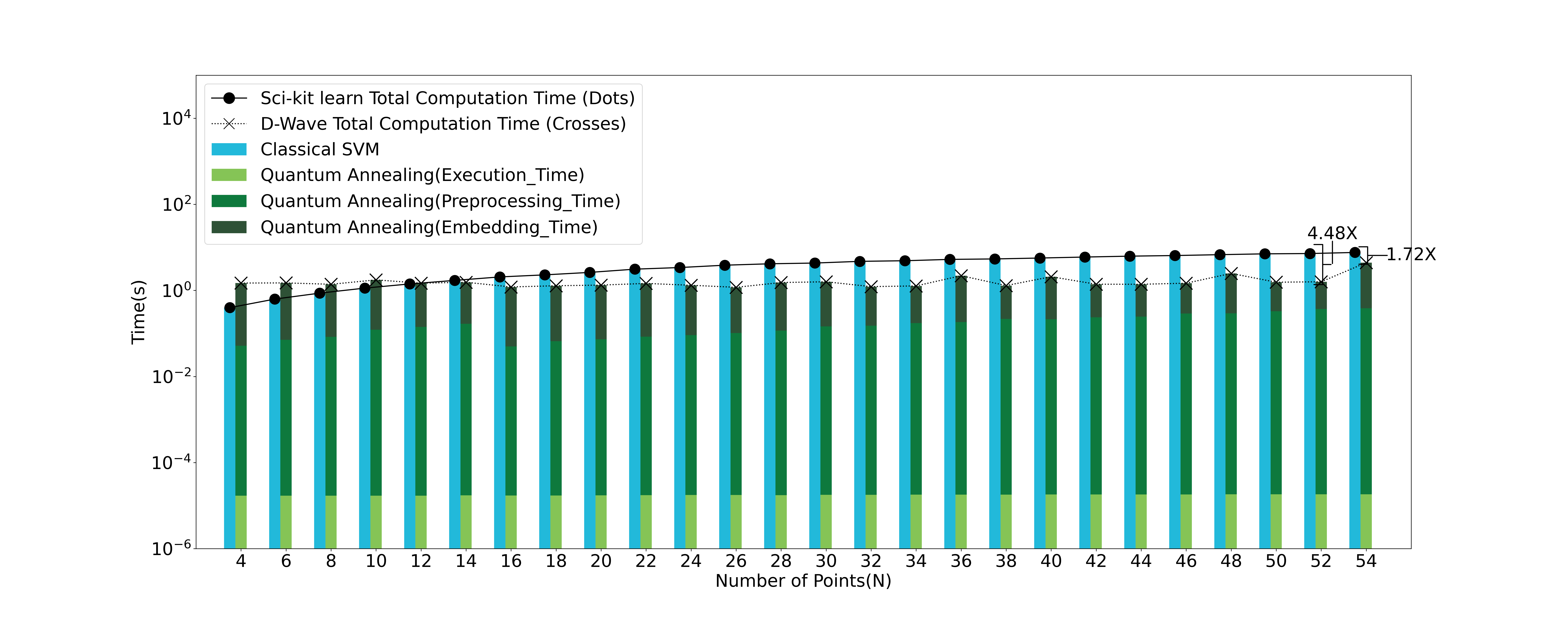}
  \caption{Scalability comparison of the Scikit-learn SVM (blue bar and bold line) and quantum SVM (light, medium, and dark green bars and dotted line). The $x$-axis indicates the number of points ($N$), and the $y$-axis logarithmically represents time in seconds. The $x$-axis ranges from $4$ to $54$. In Figure \ref{fig:numPts_time}, at 52 points, the quantum approach demonstrates a speedup of 4.48$\times$ over the classical approach. At 54 points, the quantum approach demonstrates a speedup of 1.72$\times$ over the classical approach.}
  \label{fig:numPts_time}
\end{figure*}

\subsubsection{Digits (0--1)}
The Digits dataset depicts numerals between 0 and 9 in black-and-white $8\times8$ images. Each pixel value is between 0 and 16, where 16 is the darkest. Each image has its 64 pixels represented in one row. So, in training, each point has 64 features. The value of each pixel is normalized. We train 20 points each of class 0 and class 1 for 10 iterations to observe the accuracy. To test accuracy, 158 points are used. In Table~\ref{tab:accuracy}, the classes are represented as Digits Class 0--1.
The classical approach resulted in a \(100\%\) training accuracy and a \(99.2\%\) testing accuracy. The quantum approach resulted in a \(99\%\) training accuracy and a \(97.5\%\) testing accuracy. The simulated annealer failed to find a solution for the QUBO problems.

\subsubsection{Lambeq}
Lambeq is a dataset consisting of concise three to four word sentences classified into the domains of cooking and technology, with each word labeled according to its respective part of speech (e.g., verb, noun). We train fifty-two of the seventy sentences ten times and test thirty points each time. The dataset underwent parsing and embedding by using Bert's natural language embedder. This was followed by training through an SVM model.
The classical approach resulted in a \(100\%\) training and testing accuracy. The quantum approach resulted in a \(93.5\%\) training accuracy and a \(88.7\%\) testing accuracy. The simulated annealer failed to find a solution for the QUBO problems.

\subsection{Annealing Time}
The accuracy of the quantum SVM  comes second to that of the classical SVM in certain data sets. So, we saw the need to compare the impact of the annealing time parameter, which may influence the quantum annealer's accuracy. We tested to see if the variance in annealing time has a significant effect on the accuracies of the trained models.
Each dataset is trained and tested 10 times for each condition. The data used for the experiment included the Iris dataset with the classes Versicolor and Virginica, Wisconsin Breast Cancer, and Wine dataset with the classes 1 and 2. The annealing time used for the experiment included 20$\, \mu s$, 100$\, \mu s$, and 1000$\, \mu s$. Each training set contained 52 training points. 
The results, which include the mean and standard deviation of the accuracy, are presented in Table \ref{tab:annealing_time}. 

Even as the annealing time increased, the Scikit-learn model consistently performed better and consisted less variance in accuracy than the quantum and simulated annealing models.
There wasn't a statistically significant difference in the accuracy by increasing the annealing time as the mean accuracies of each test were within one standard deviation of each other. However, the Iris data set with the classes Virginica and Versicolor saw less variance and higher training and testing accuracy when trained for $1,000 \mu s$ as  compared to $20 \mu s$.

\subsection{Time}
\label{sub:scalability-results}
We conduct a scalability analysis of the training time of the classical and quantum approach using synthetic datasets. We generate the random datasets using the Scikit-learn \texttt{make\_blob} function. Before training the SVM models, we normalize the data points. The D-Wave access time represents the programming and sampling time. The sampling time includes annealing, read out, delay time.  
The simulated annealing approach failed to find a solution when there are many features, so we decided it was inappropriate to examine the scalability of the simulated annealing approach.
The complexity of the classical approach is O($N^3$), and the complexity of the quantum approach is O($N^2$). 

\subsubsection{Number of Features} 
We conduct a scalability study to investigate the impact of feature size on the compute time of our classical and quantum SVM approaches. We also describe the runtime performance as the number of features increase from 2 to 8,388,608. The number of data points used for training remains constant at 52 across all the datasets because that is the most data points we could stably embed on the D-Wave Advantage quantum computer. 
The number of qubits available limits the number of data points. We are also testing each configuration 10 times.

The results, including the mean and standard deviation obtained from 10 iterations, are presented in Table~\ref{tab:features}. The scalability is shown in Figure~\ref{fig:svm-scaling-d}, where the $x$-axis indicates the number of features ($d$), and the $y$-axis logarithmically represents time in seconds. We depict the total Scikit-learn time with light blue bars. The medium green bars represent the D-Wave preprocessing time. The light green bars represent the D-Wave access time, and the dark green bars show the D-Wave embedding time. As shown, the access time stays constant at 18 ms. 
With smaller number of features ($d \leq$ 1,048,576), the classical approach's runtime is faster than the D-Wave approach. As the number of features increases ($d \geq$ 2,098,152), the quantum approach's compute time is faster than the classical approach. With fewer features, the time taken for embedding greatly influences the overall compute time on the D-Wave system. 
In contrast, the difference in access and preprocessing time has less influence. Similarly, with many features, the embedding time remains the dominant factor in the D-Wave's compute time, with the access and preprocessing times have minimal impact. However, the preprocessing time does increase, whereas the embedding and access time do not. The performance of both approaches is comparable when there are 2 million features. When training with up to 8 million features, the quantum approach exhibits a speedup of 3.69$\times$ over the classical approach. Although not considered for the total computation time, the network overhead time increases as the number of points increases. The increased points result in an increased number of qubits used, which means longer data transfer times to the D-Wave server. The testing and training accuracies are both \(100\%\).

\subsubsection{Number of Data Points}
We also conduct a scalability study to investigate how the number of data points affects the compute time of our classical and quantum SVM approaches. We examine the runtime performance as the dataset expands from 4 to 54 data points. The number of features used for training remains constant at 8 million across all the datasets because the runtime performance of both approaches is comparable at 12 points and 8 million features. The number of qubits available limits the number of points. 

The results, including the mean and standard deviation obtained from 10 iterations, are presented in Table~\ref{tab:trained-points}. We present the scalability results in Figure~\ref{fig:numPts_time}, where the $x$-axis indicates the number of data points ($N$), and the $y$-axis logarithmically represents time in seconds. We depict the total Scikit-learn time with blue bars. The green bars represent the D-Wave preprocessing time. The light green bars represent the D-Wave access time, and the dark green bars show the D-Wave embedding time. The dominant factor in the total time taken for the Quantum SVM is the embedding time. The preprocessing and access time increase as the number of points increases, while the embedding time remains relatively constant, except when training 54 points. Considering the available number of qubits, this outcome is expected as the data size increases. However, even at 54 points, the quantum SVM is 1.72$\times$ faster than its classical counterpart. At 52 points, where the quantum computer could find an embedding with minimal time fluctuation, the quantum computer outperforms the classical computer by a factor of 4.48.

\subsection{Discussion}
\label{sub:discussion}
To the best of our knowledge, this is one of the first studies that shows a demonstrable quantum speedup for training machine learning models on the NISQ-era quantum computers. 
Our study, coupled with the linear regression study by Date and Potok in 2021 \cite{date2021adiabatic}, provides a compelling evidence that a quantum speedup can be realized on datasets that have either a large number of data points or a large number of features. 
On smaller datasets, the overheads of running jobs on the quantum computer overshadow any speedup obtained by the quantum approach.

The testing accuracies presented in Table \ref{tab:accuracy} show that the quantum approach performs better than the classical approach in case of Wine 1--2.
On some datasets such as synthetic and Iris, it equals the performance of the classical approach.
On other datasets however (Iris Versicolor-Virginica, WBC, Wine 0--2, Digits, and Lambeq), it is outperformed by the classical approach.
On these datasets, the lower accuracy of the quantum approach is partly attributed to the fact that the datasets were linearly inseparable, for e.g., Versicolor-Virginica.
For other datasets, the accuracy of the weights learned in the quantum approach was directly affected by the numerical precision (i.e., number of bits) allocated to each Lagrangian multiplier in the QUBO problem.
This precision was in turn constrained by the hardware architecture, i.e., the number of qubits and their connectivity available on the quantum hardware.
Within the constraints imposed by the quantum architecture, the quantum approach was able to produce accuracies that were in the same ballpark as the classical approach for most of the datasets.

Our results in this paper demonstrate that the quantum approach trains SVMs faster than the Scikit-learn classical approach on larger-sized datasets.
While the Scikit-learn's algorithm for training SVMs may not be the most efficient classical algorithm, it certainly is one of the most widely used algorithms in the literature and runs in $\mathcal{O}(N^3)$ time.
The classical algorithms have been optimized for decades, whereas the quantum algorithms are still in their nascent stages. 
In this light, our motivation in this paper was to demonstrate that our quantum approach can perform faster than a widely used classical approach.
Our results are intended to serve as a baseline for more optimized quantum approaches in the future.
We certainly hope that the future quantum approaches would outperform not only the most widely used classical approaches, but also the \emph{best} performing classical approaches.

% may not outperform the \emph{best} classical approach yet. Our motivation was to demonstrate that it performs better than \emph{a} classical approach. The Scikit-learn SVM is a machine learning model that is widely used within the scientific and engineering community and serves as a good baseline for comparison.

\section{Conclusion}
A critical limitation of current state-of-the-art machine learning is the extensive computational resources required for training. The duration of this process varies significantly---ranging from a few hours to several months---and is contingent upon the
size of the training dataset. To address this problem, we introduce a quantum approach for solving the SVM problem. This paper describes our empirical analysis of transforming SVMs to QUBO problems, which D-Wave Advantage computers can solve. Through theoretical analysis, we establish that our quantum approach outperforms the current classical approach for runtime. Next, we compare the performance of an SVM implementation on a D-Wave Advantage quantum annealer with a classical implementation on AMD Ryzen 5 4600H and Intel Xeon E5-2690v4 CPUs. Our training results demonstrate that the quantum approach's accuracy is comparable with the classical approach. When training with a dataset with 8 million features, the quantum annealer outperforms the classical computer by achieving up to 4.48$\times$ faster computation. Our research contribution shows that quantum computing can effectively reduce training times and lead to accelerated scientific discoveries and improved performance for machine learning models.

In the future, we would like to explore novel approaches to train larger datasets effectively. We want to expand our quantum approach to encompass variants of SVMs that leverage kernel methods. Finally, we are interested in exploring techniques to mitigate the effects of noise and errors in quantum annealing.
\label{sec:conclusion}

\section{Acknowledgment}
\label{sec:ack}
This manuscript has been authored by UT-Battelle LLC under contract DE-AC05-00OR22725 with the US Department of Energy (DOE). The US government retains and the publisher, by accepting the article for publication, acknowledges that the US government retains a nonexclusive, paid-up, irrevocable, worldwide license to publish or reproduce the published form of this manuscript, or allow others to do so, for US government purposes. DOE will provide public access to these results of federally sponsored research in accordance with the DOE Public Access Plan (\url{https://www.energy.gov/doe-public-access-plan}).
This research was supported by the Exascale Computing Project (17-SC-20-SC), a collaborative effort of the U.S. Department of Energy Office of Science and the National Nuclear Security Administration.
The authors would like to thank Sam Crawford of Oak Ridge National Laboratory for editing this manuscript.

\bibliography{example_paper}
\bibliographystyle{icml2022}

%%%%%%%%%%%%%%%%%%%%%%%%%%%%%%%%%%%%%%%%%%%%%%%%%%%%%%%%%%%%%%%%%%%%%%%%%%%%%%%
%%%%%%%%%%%%%%%%%%%%%%%%%%%%%%%%%%%%%%%%%%%%%%%%%%%%%%%%%%%%%%%%%%%%%%%%%%%%%%%
% APPENDIX
%%%%%%%%%%%%%%%%%%%%%%%%%%%%%%%%%%%%%%%%%%%%%%%%%%%%%%%%%%%%%%%%%%%%%%%%%%%%%%%
%%%%%%%%%%%%%%%%%%%%%%%%%%%%%%%%%%%%%%%%%%%%%%%%%%%%%%%%%%%%%%%%%%%%%%%%%%%%%%%
% \newpage
% \appendix
% \onecolumn

% You can have as much text here as you want. The main body must be at most $8$ pages long.
% For the final version, one more page can be added.
% If you want, you can use an appendix like this one, even using the one-column format.
%%%%%%%%%%%%%%%%%%%%%%%%%%%%%%%%%%%%%%%%%%%%%%%%%%%%%%%%%%%%%%%%%%%%%%%%%%%%%%%
%%%%%%%%%%%%%%%%%%%%%%%%%%%%%%%%%%%%%%%%%%%%%%%%%%%%%%%%%%%%%%%%%%%%%%%%%%%%%%%

\end{document}